# Contrastive Clustering: Toward Unsupervised Bias Reduction for Emotion and Sentiment Classification

Jared Mowery
The MITRE Corporation

## Abstract

**Background:** When neural network emotion and sentiment classifiers are used in public health informatics studies, biases present in the classifiers could produce inadvertently misleading results.

**Objective:** This study assesses the impact of bias on COVID-19 topics, and demonstrates an automatic algorithm for reducing bias when applied to COVID-19 social media texts. This could help public health informatics studies produce more timely results during crises, with a reduced risk of misleading results.

**Methods:** Emotion and sentiment classifiers were applied to COVID-19 data before and after debiasing the classifiers using unsupervised contrastive clustering. Contrastive clustering approximates the degree to which tokens exhibit a causal versus correlational relationship with emotion or sentiment, by contrasting the tokens' relative salience to topics versus emotions or sentiments.

**Results:** Contrastive clustering distinguishes correlation from causation for tokens with an $F_1$ score of 0.753. Masking bias prone tokens from the classifier input decreases the classifier's overall $F_1$ score by 0.02 (anger) and 0.033 (negative sentiment), but improves the $F_1$ score for sentences annotated as bias prone by 0.155 (anger) and 0.103 (negative sentiment). Averaging across topics, debiasing reduces anger estimates by 14.4% and negative sentiment estimates by 8.0%.

**Conclusions:** Contrastive clustering reduces algorithmic bias in emotion and sentiment classification for social media text pertaining to the COVID-19 pandemic. Public health informatics studies should account for bias, due to its prevalence across a range of topics. Further research is needed to improve bias reduction techniques and to explore the adverse impact of bias on public health informatics analyses.

**Correspondence:** jmowery@mitre.org

# Introduction

The volume and timeliness of social media data has led to its increasingly widespread use in public health informatics [1], including the increasing use of machine learning techniques to efficiently analyze that data [2]. The COVID-19 pandemic highlights the need for rapid and robust public health informatics analyses to guide policymakers during crises. Public health informatics studies using social media data have employed sentiment analysis algorithms to address many aspects of the pandemic, including studying the "infodemic" of online discussions [3], public perceptions of social distancing [4], vaccine hesitancy and disinformation [5], and the mental health impact of the pandemic [6]. However, neural network classifiers develop bias based on their training data across a wide range of applications. These biases could impact the accuracy of public health informatics analyses, potentially inadvertently misleading policymakers. Fortunately, researchers have been developing algorithms to make the classifiers less biased [7], which may reduce the adverse effect of bias on public health informatics analyses.

This study focuses on bias in emotion and sentiment classification for COVID-19 social media data, which could result in overestimates or underestimates of the public health phenomena being studied. Other sources of bias, such as self-selection biases, inaccuracies in automated account ("bot") activity detection in social media [8], or potential confirmation biases [9], are outside the scope of this study.

Supervised approaches, which use annotated data, have been developed to counteract classifier bias using prior information. For example, adversarial learning approaches have been developed to use author demographic information to counter bias and ensure fairer representations [10] [11], while annotations of words for their relevance to the classification task have been used to train neural networks to reduce the bias they learn from their training data [12]. Additionally, modifying a classifier's input to mask input words annotated as inducing bias can reduce classifier bias without retraining the classifier [13]. Most neural network text classifiers use neural language models or word embeddings, which represent the structure of language and the latent semantic relationships between words, but these representations have been shown to incorporate bias [14]. However, prior knowledge can be used to mitigate the bias in these representations [15] [16]. Studies may also examine multiple techniques, such as exploring debiased word embeddings, swapping gender-specific words in training data, and using larger corpora [17].

Supervised approaches benefit from accurately annotated data or carefully engineered features (e.g. swapping gender-specific words), but annotating the data or devising the feature engineering approaches can be time-consuming, and certain types of prior information, such as author demographic attributes, may raise ethical, legal, or privacy concerns.

Unsupervised or minimally supervised approaches seek to minimize or eliminate the amount of annotated training data or feature engineering required to reduce bias. Many approaches leverage words referring to marginalized or disadvantaged groups to reduce bias with a minimum of hand annotation, such as leveraging gendered pronouns to reduce gender bias. For example, gender bias can be reduced in predicting occupations from people's biographies by leveraging biases present in word embeddings [18], and gender bias in literature can be explored by contrasting topics built using gendered words with non-gendered topics [19]. Other approaches are more general purpose, such as hypothesizing that protected demographic variables will often correlate with known variables (e.g. location) or with sparse regions of training data (since training data for underrepresented groups is often rarer). These regions can be identified via poor classifier accuracy or data sparsity with little or no prior information, and they can be leveraged to guide the classifier's training to improve accuracy for these regions [20] [21] [22].

Although unsupervised or minimally supervised approaches are less time-consuming and may avoid ethical, legal or privacy concerns, their effectiveness in reducing bias may be hindered by the lack of human decision-making in their training data and features.

Classifier bias has been demonstrated in sentiment analysis systems [23], which raises questions about the impact of sentiment analysis-related biases. The need for rapid public health informatics analyses of crises like the COVID-19 pandemic raises an additional challenge: the sources of bias can change quickly and unexpectedly. Words like "hydroxychloroquine" can suddenly become controversial and serve as potential sources of bias. Common words like "virus", that are traditionally associated with negative sentiments or emotions, may introduce bias that goes unnoticed in other applications, but could lead to overestimates of anger or negative sentiment in studies of COVID-19. These unexpected sources of bias present a challenge for both supervised and unsupervised approaches to bias reduction: supervised approaches are ill-suited to rapid analyses due to requiring annotated data, while unsupervised approaches will often miss unexpected sources of bias arising from COVID-19 due to their focus on biases related to demographics.

This study addresses the challenges described above for reducing bias in COVID-19 analyses by combining and extending prior research from supervised and unsupervised approaches.

## Methods

Prior research has demonstrated that bias can be reduced by masking words in a classifier's input that belong to an annotated list of bias prone words. The method in this study eliminates the time-consuming tasks of annotating data or re-training classifiers, while still achieving a reduction in bias, by applying contrastive clustering as an unsupervised method to automatically identify bias prone tokens (e.g. words, hashtags, URLs, emoticons, emojis), and then mask those tokens when applying emotion and sentiment classifiers.

The approach taken in this paper can be divided into four phases: data gathering and preparation (Data Gathering and Preparation), contrastive clustering analysis to identify bias prone tokens (Contrastive Clustering), evaluation of the accuracy of the contrastive clustering algorithm against human judgment using token-level annotations (Token-Level Annotation), and bias mitigation of the emotion and sentiment classifier using the bias prone tokens (Bias Mitigation).

**Data Gathering and Preparation**

The data gathering and preparation phase involves gathering the COVID-19 related social media data used in this study, hand annotating it for testing purposes, and applying three machine learning algorithms to prepare the data for contrastive clustering. The contrastive clustering algorithm does not use the hand annotations—the sentences were only hand-annotated for the purpose of testing the accuracy of the original and debiased versions of the emotion and sentiment classifiers, and assessing the impact of bias for COVID-19 public health informatics studies. The annotations were also not used for the three machine learning algorithms, since they are either unsupervised or were already trained on non-COVID-19 data.

The data used in this study was gathered from Reddit, a social media discussion forum, between November 2019 and June 2020. Since the posts or comments can vary greatly in length, sentence segmentation was used to extract the sentences from each post or comment. A set of 10,000 sentences were selected which contained keywords and phrases referring to COVID-19, including social distancing, quarantine measures and political or policy discussions.

To support later testing and evaluation, the annotators labeled each of the 10,000 sentences for whether they exhibited each of the following emotions: anger, disgust, anticipation, love, sadness, joy, and fear. A sentence could be marked with multiple emotions, such as anger and disgust. If the sentence contained no apparent emotions, it was marked as "none". If the sentence was ambiguous, the annotators could choose to skip it. In addition, the annotators were asked to indicate when a sentence did not exhibit an emotion, but might appear to exhibit the emotion *to a biased observer.* For example, if a sentence states that most people were not wearing masks in a situation, and the sentence contains no obvious indications of emotion, the author may be expressing anger, but the author may also simply be making an observation. Since it is not clear whether the author is expressing emotion, an unbiased observer would either mark the sentence as "none" or skip it, but a biased observer would assume that it should be marked "anger", reflecting not necessarily the author's emotion, but the emotion of the biased observer.

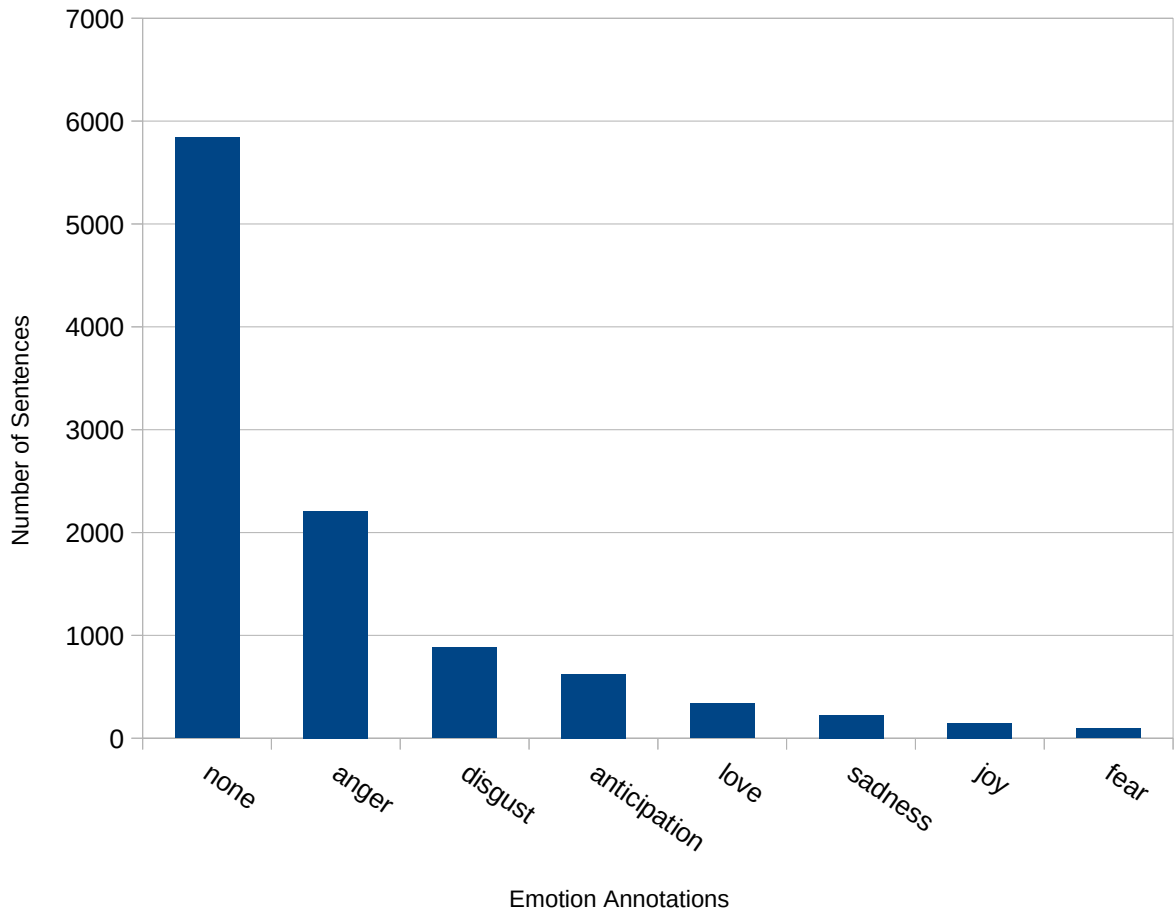

Figure 1: Prevalence of Emotions in Annotation Data. Each of 9,169 sentences was annotated for the emotions shown. The majority of sentences contained no emotion, followed by anger (2,210) and disgust (883). Sentences can have multiple emotion labels, such as anger and disgust.

This resulted in 9,169 annotated sentences (Figure 1). Of these, the majority (5,841) contained no emotion. The most prevalent emotion was anger, with 2,210 sentences, followed by disgust, with 883 sentences. There were 199 sentences marked as bias prone for anger, 9 for love, 3 for disgust, and one each for joy, sadness, and fear. This study only analyzed potential bias for anger, since the other emotions contained too few examples.

Independently of the hand annotation task, three machine learning algorithms were applied to automatically label the set of 10,000 sentences: an emotion classifier, a sentiment classifier, and a topic modeling algorithm. Since the sentences only contain significant amounts of anger, only the anger label was used from the emotion classifier and only the negative sentiment label was used from the sentiment classifier. All labels produced by the topic modeling algorithm were used. This results in each sentence being labeled with inferences for anger, negative sentiment, and a set of topics.

The pre-trained sentiment and emotion classifiers are from the TweetEval benchmark study [24]. These classifiers were originally trained for Twitter data, which consists of relatively short text; applying sentence segmentation to the Reddit posts and comments produces shorter text that is more suitable for these classifiers. The third algorithm, BERTopic, automatically detects topics present in each sentence [25]. BERTopic was set to produce 20 topics, which resulted in a reasonable variety of coherent topics. Any sentence which had no topic confidence values above 0.1 for any of the 20 topics was assigned to a 21$^{st}$ "topic" representing no discernible topic.

**Contrastive Clustering**

The contrastive clustering algorithm is designed to reduce bias by distinguishing whether tokens (such as words, names, punctuation marks, emoticons, emojis, numbers or URLs) constitute expressions of emotion, or merely correlate with expressions of emotion. Tokens that appear to correlate with emotions or sentiment but are not expressions of emotion or sentiment are *distractors*. The underlying hypothesis is that algorithmic bias is created partly because neural network classifiers cannot distinguish correlation from causation, so they cannot recognize distractors, and consequently may treat distractors as indications of genuine expressions of emotion or sentiment.

Anecdotally, distractors are frequently the *targets* of expressions of emotion, and relate closely to the topic of conversation. Consequently, contrastive clustering examines how closely words or tokens correlate with sentiments or emotions (via the TweetEval pre-trained sentiment and emotion classifiers) versus topics (via BERTopic). If the word or token is more strongly associated with a topic than an emotion or sentiment, then it is probably a distractor. For example, the name of a controversial policy might frequently occur in sentences in which people express anger, making it a distractor. If the policy name correlates with sentiment or emotion sufficiently strongly, then the sentiment or emotion classifiers are likely to mistakenly learn that the policy name itself denotes anger, which will result in bias. Due to this bias, the classifier will be more likely to classify sentences containing the policy name as expressions of anger, regardless of whether the sentence actually contains any expressions of anger.

The remainder of this section defines the contrastive clustering algorithm. First, the vocabulary of tokens *T* contains the 10,000 most frequently occurring tokens across all sentences *S*. The remaining tokens were ignored for the contrastive clustering algorithm, so no token outside of *T* will be masked. The selection of the vocabulary *T* has no effect on the TweetEvel classifiers and BERTopic, since those algorithms define their own vocabularies.

Now, each subset of sentences in S corresponding to the TweetEval classifiers' determinations of sentiment or emotion, or to BERTopic's assignments of topics to sentences, is a group. Since anger was the only prevalent emotion from the hand annotated data, only the TweetEval classifiers' anger and negative sentiment results are useful, resulting in the following groups:

$$G_{emotion} = \{G_{anger}, G_{negative}\}$$

$$G_{topic} = \{G_{t1}, G_{t2}, G_{t3}, ... G_{t21}\}$$

$$G = G_{emotion} \cup G_{topic}$$

Next, the probability that a sentence will belong to group $g_i \in G$ given that the sentence contains token $t_j \in T$ is:

$$P(g_i|t_j) = WilsonCC_{Lower}(count(t_j, g_i), num(t_j))$$

where *WilsonCC$_{Lower}$* is the lower bound of the Wilson confidence interval with continuity correction ([26]), *count(t$_j$, g$_i$)* is the number of sentences in which token *t$_j$* was present and the sentence belonged to group *g$_i$*, and *num(t$_j$)* is the number of sentences that contained token *t$_j$*. Using the lower bound of the confidence interval reduces the risk that tokens will falsely appear to be strongly associated with groups, when they merely occur infrequently and they only correlated with the group due to luck. For example, if a token only occurs twice in the corpus, and both times it appeared in sentences classified by the TweetEval classifier as anger, then the token would get a probability of 1 for being an expression of anger. However, that reflects too much confidence for a token that only appears twice in the corpus. Since the lower bound takes the number of times the token has been observed into account, it produces a considerably lower, more conservative estimate of the probability. The maximum conditionally independent probabilities that the presence of a token *t$_j$* indicates the sentence will belong to either the set of emotion groups or the set of topic groups are defined as:

$$P(G_{emotion}|t_j) = \max_i P(g_i|t_j), g_i \in G_{emotion}$$

$$P(G_{topic}|t_j) = \max_i P(g_i|t_j), g_i \in G_{topic}$$

Intuitively, token *t$_j$* is likely to be a distractor when $P(G_{topic}|t_j) > P(G_{emotion}|t_j)$. However, while the lower bound equation takes the frequency of occurrence of the tokens into account, it does not account for the differing distributions of tokens and groups. Define the distribution for a group *g$_i$* as:

$$D_i = \underset{j}{sort}\ P(g_i|t_j)$$

where the sort function sorts the probabilities in descending order. Figures 2 and 3 illustrate the distributions for the anger emotion group and the ventilators topic group. The shapes of these distributions are significantly different. Other distributions, especially for less common topics, have considerably lower probability values even for their most salient tokens. To partly compensate, the contrastive clustering algorithm includes a coefficient to adjust the relative importance of topic and emotion probabilities:

$$C_{expressive}(t_j) = P(G_{emotion}|t_j)/2 + 0.5, when\ P(G_{emotion}|t_j) \geq \alpha P(G_{topic}|t_j)$$

$$C_{expressive}(t_j) = -P(G_{topic}|t_j)/2 + 0.5, when\ P(G_{emotion}|t_j) < \alpha P(G_{topic}|t_j)$$

where $C_{expressive}(t_j)$ denotes the confidence estimate that token $t_j$ expresses an emotion, rather than being a distractor.

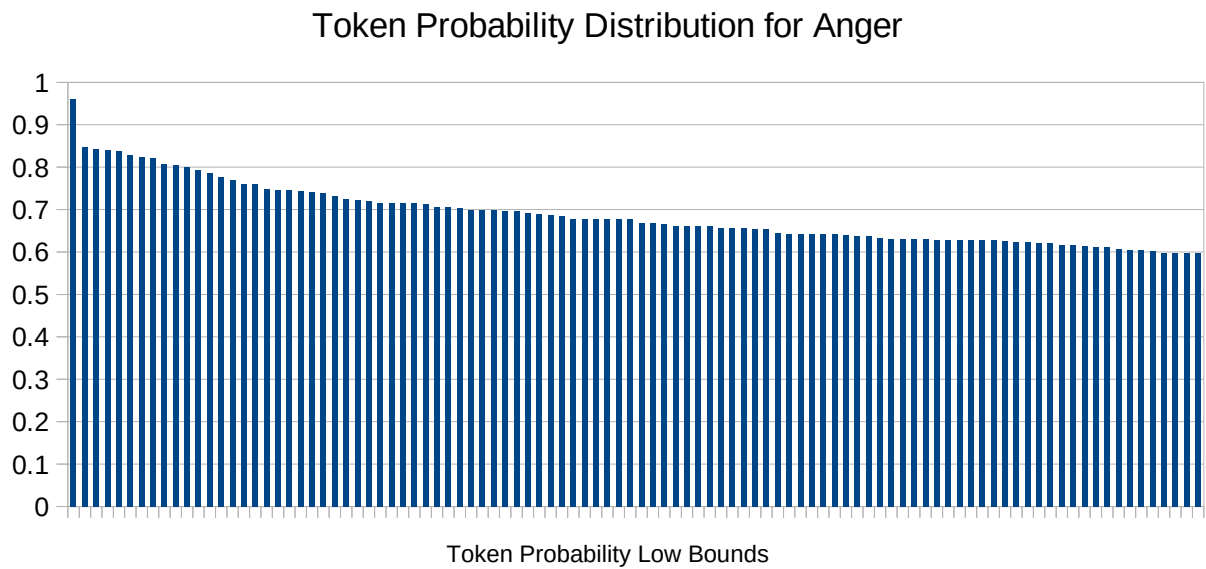

Figure 2: The distribution of token probability lower bounds for the anger group, for the 100 highest probability tokens. The probability values are high and the distribution is relatively flat, indicating a large number of relevant tokens. (X-axis token labels omitted due to frequent profanity.)

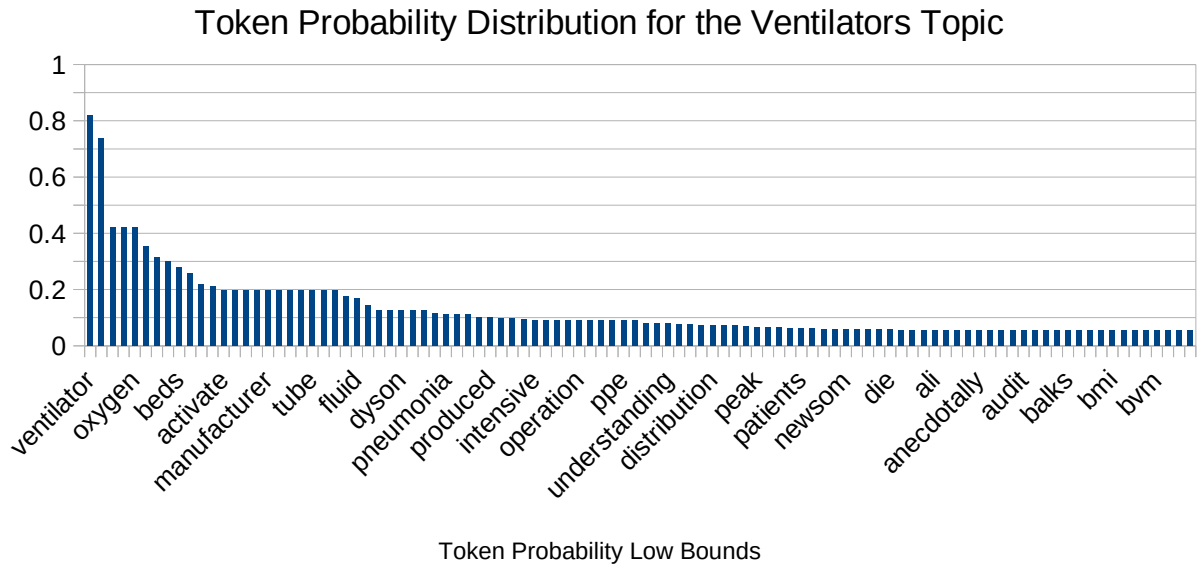

Figure 3: The distribution of token probability lower bounds for a topic focused on ventilators, for the 100 highest probability tokens. The probability values sharply decline and the distribution has a long tail, indicating only a few highly salient tokens.

The Results section will include testing of the contrastive clustering algorithm based on the $C_{expressive}(t_j)$ confidence values. However, contrasting the topic and emotion groups only provides a partial solution to the correlation versus causation problem. There are many reasons a token may appear to be expressive even when it is not, or vice versa, including noise (e.g. tokens that appear related to an emotion or sentiment by luck), linguistic and grammatical patterns that prevent tokens from being independently and identically distributed in a manner that correlates with emotions or topics, differences in token connotations between the TweetEval classifiers' pre-COVID training data and the COVID-19 data, and optimistic conditional independence assumptions or simplifications in the implementation of the contrastive clustering algorithm itself (e.g. using a topic coefficient to account for differences in distributions rather than a more complex, parametric statistical language model).

To reduce these errors, the final step of the contrastive clustering algorithm examines tokens that are semantically related to identify tokens inferred as expressive whose semantic nearest neighbors are non-expressive, or vice versa. These tokens are likely to have been incorrectly labeled by the contrastive clustering algorithm, since semantically similar tokens will typically all be expressive (e.g. profanity) or non-expressive (e.g. topic-related tokens). To quantify semantic similarity, this study uses FastText word embeddings [27]. Word embeddings are produced using neural networks to represent words as vectors based on each word's contexts in a large corpus (e.g. nearby tokens in each sentence). This results in the proximity of two words' vectors in the embedding space approximating the semantic similarity of the words. An approximate nearest neighbors algorithm ([28]) was used to efficiently identify semantically similar tokens.

For each token $t_j$, the nearest neighbors algorithm was used to identify the nearest $n$ neighbors and their confidence scores for being expressive tokens, including $t_j$ itself:

$$N(t_j) = \{C_{expressive}(t_1), C_{expressive}(t_2), C_{expressive}(t_3), \ldots C_{expressive}(t_n)\}$$

These neighbors were sorted and then the *k*-th nearest neighbor was selected based on:

$$k = round(\beta \cdot |N(t_j) - 1|)$$

where $\beta$ is a coefficient used to control the significance of the nearest neighbors when considering relabeling $t_j$. The expressiveness confidence score for the *k*-th nearest neighbor was used to replace the original expressiveness confidence score for token $t_j$:

$$\bar{C}_{expressive}(t_j) = N(t_j)[k]$$

This has the effect of "smoothing" the confidence scores across all tokens, so neighboring tokens are more likely to share the same confidence score.

**Token-Level Annotation**

The contrastive clustering algorithm produces confidence estimates that each token is an expression of an emotion rather than a distractor. To test the accuracy of the algorithm, those confidence estimates were compared to human judgment for the 1,000 tokens with the highest association with each of the anger and negative sentiment groups, based on the lower bound of the Wilson confidence interval with continuity correction for each group. Each token was marked as expressive if it was a direct expression of anger or negative sentiment (e.g. an obscenity), an insult (e.g. "liar", "incompetent", or "crazy"), or carried a negative connotation (e.g. writing "claimed" to express distrust rather than "said" or "stated"). The token was marked as non-expressive (a distractor) if it pertained to a typical target of emotion but did not itself convey an emotion (e.g. "quarantine", "virus", "impeachment", or names of public figures), or if the token was irrelevant or too ambiguous for emotion or sentiment classification (e.g. "it", "the", "next", "today"). Since the tokens correlated with anger and negative sentiment frequently overlap, of the 2,000 tokens annotated, 1,339 were unique. These included 209 tokens labeled as expressive and 1,130 labeled as distractors. The distractor tokens consisted of 318 tokens representing typical targets of emotion and 812 irrelevant or ambiguous tokens. The Results section presents the results for testing the contrastive clustering algorithm on its ability to distinguish the expressive tokens from distractor tokens.

**Bias Mitigation**

To reduce bias in the emotion and sentiment classifiers, the expressiveness scores from the contrastive clustering algorithm are used to "mask" likely distractor tokens from the classifier's input. The tokens are masked by replacing them with the word "it" (as a neutral pronoun), so that the classifier's decision is not influenced by the masked token. To determine which tokens are masked, the algorithm considers the top *N* tokens with the highest values for the lower bound of the Wilson confidence interval with continuity correction, using a group defined to include sentences classified as containing either anger or negative sentiment by the classifier. The emotion and sentiment classifiers are most likely to learn that these tokens indicate expressions of anger or negative sentiment, even if the tokens are distractors. Next, for each token *t*, the algorithm chooses to mask *t* when:

$$\bar{C}_{expressive}(t) < 0.5$$

## Results

This section provides results for evaluating the accuracy of the contrastive clustering algorithm's per-token confidence estimates, the extent of bias reduction in the TweetEval classifiers when masking tokens, and a per-topic examination of how debiasing the classifiers affects estimates of anger and negative sentiment prevalence that would impact public health informatics analyses.

### Per-Token Accuracy

Evaluating per-token accuracy provides an indication of whether the contrastive clustering algorithm's token confidence values help mitigate the classic correlation versus causation problem.

The confidence estimates based on the statistical analysis—without applying smoothing via nearest neighbor analyses of word embeddings—are denoted the Expressive Confidences ( $C_{expressive}(t_j)$ ), while the versions with smoothing are denoted the Smoothed Confidences ( $\bar{C}_{expressive}(t_j)$ ). In each case, the $F_1$ scores are calculated by comparing the confidence values to the hand annotations for each token. A threshold value is used to determine when the confidence is high enough to denote the token as expressive; higher threshold values favor parsimoniously designating tokens as expressive.

Table 1: Expressive confidence F1 scores as a function of the topic coefficient α and the threshold above which the token is assumed to be expressive. The topic coefficient's default value of 1, which gives equal weight to a token's apparent affinity for topic groups and for emotion groups, yields the best result. The default threshold of 0.5 produces a sub-optimal F1, likely due to expressive tokens being rarer than distractor tokens in the data set. A threshold of 0.8 produces the optimal F1 score: 0.672.

| Topic Coefficient α | Threshold | | | | |
|---|---|---|---|---|---|
|  | 0.5 | 0.6 | 0.7 | 0.8 | 0.9 |
| 0 | 0.14 | 0.14 | 0.44 | 0.66 | 0.5 |
| 0.5 | 0.17 | 0.17 | 0.44 | 0.66 | 0.5 |
| 1 | 0.62 | 0.62 | 0.65 | **0.67** | 0.5 |
| 1.5 | 0.59 | 0.59 | 0.58 | 0.53 | 0.46 |
| 2 | 0.53 | 0.53 | 0.51 | 0.48 | 0.46 |

The highest macro-averaged $F_1$ score for the Expressive Confidences set when choosing the optimal topic coefficient and threshold is 0.672, while using a naive threshold setting yields only 0.617 (Table 1).

Table 2: Smoothed confidence F1 scores as a function of the number of nearest neighbors and the median coefficient β.

| Median Coefficient β | Number of nearest neighbors | | | | | | | |
|---|---|---|---|---|---|---|---|---|
| | 2 | 3 | 4 | 5 | 6 | 7 | 8 | 9 |
| 0.1 | 0.69 | 0.71 | 0.71 | 0.68 | 0.66 | 0.72 | 0.73 | 0.72 |
| 0.2 | 0.69 | 0.71 | 0.71 | 0.72 | 0.72 | 0.72 | 0.73 | **0.75** |
| 0.3 | 0.69 | 0.65 | 0.71 | 0.72 | 0.71 | 0.74 | 0.75 | 0.75 |
| 0.4 | 0.69 | 0.65 | 0.71 | 0.67 | 0.71 | 0.74 | 0.72 | 0.73 |

The highest macro-averaged $F_1$ score for the Smoothed Confidences was 0.753, which occurred when using the maximum number of nearest neighbors (9), the default threshold setting of 0.5, and a median coefficient β of 0.2 (Table 2). For 2 through 9 nearest neighbors, the optimal $F_1$ scores occurred with the default threshold value of 0.5 and with median coefficients near 0.2 or 0.3. Consequently, for Smoothed Confidences, the median coefficient serves a similar purpose as the threshold for sparingly labeling tokens as expressive.

**Classifier Debiasing Accuracy**

Table 3: Macro-averaged F1 scores for the original and debiased versions of the anger classifier, as a function of the number of tokens considered for masking and sets of sentences reflecting varying degrees of bias. The biased sentence sets include all sentences, sentences exceeding bias propensity estimates, and bias prone sentences determined via hand annotation. For each bias propensity value, sentences were included only if their bias propensity score met or exceeded the listed value. For each set of sentences, the n value denotes the number of sentences included.

| Anger | | Biased Sentence Sets | | | | | | |
|---|---|---|---|---|---|---|---|---|
| # Tokens | Classifier | All n = 9169 | 0.1 n = 3324 | 0.2 n = 1195 | 0.3 n = 375 | 0.4 n = 131 | 0.5 n = 56 | Bias Prone n = 199 |
| N/A | Original | **0.672** | 0.594 | 0.564 | 0.526 | 0.499 | 0.429 | 0.292 |
| 1000 | Debiased | 0.646 | 0.592 | 0.593 | 0.586 | **0.597** | 0.500 | 0.436 |
| 2000 | | 0.650 | 0.593 | **0.602** | 0.598 | 0.569 | 0.500 | 0.446 |
| 5000 | | 0.653 | 0.594 | 0.600 | **0.629** | 0.557 | 0.500 | 0.452 |
| 10000 | | 0.658 | **0.596** | 0.593 | 0.618 | 0.544 | **0.510** | **0.453** |
| Average Change | | -0.020 | 0.000 | +0.033 | +0.082 | +0.068 | +0.074 | +0.155 |

The debiased version of the anger classifier, which masks tokens based on the optimal parameter settings for the Smoothed Confidences results, consistently improves the $F_1$ score for bias prone sentences (Table 3). When all sentences are included (n = 9169), the original anger classifier achieves a better $F_1$ score by about 0.02 points. However, when the bias propensity estimate is used to focus on subsets of sentences prone to bias, the debiased version of the classifier consistently outperforms the original version, with the gap in $F_1$ scores increasing as the degree of likely bias increases. The gap is largest when using the subset of sentences that were hand annotated as bias prone, suggesting a reasonably strong connection between algorithmic bias and human perceptions of bias.

Table 4: Macro-averaged F1 scores for the original and debiased versions of the negative sentiment classifier. The negative sentiment classifier benefits less from debiasing than the anger classifier, but the patterns in the F1 scores are similar: the overall F1 is decreased in exchange for reducing bias.

| Negative Sentiment | | Biased Sentence Sets | | | | | | |
|---|---|---|---|---|---|---|---|---|
| # Tokens | Classifier | All n = 9169 | 0.1 n = 3324 | 0.2 n = 1195 | 0.3 n = 375 | 0.4 n = 131 | 0.5 n = 56 | Bias Prone n = 199 |
| N/A | Original | **0.702** | **0.657** | 0.625 | 0.581 | 0.573 | 0.524 | 0.361 |
| 1000 | Debiased | 0.670 | 0.602 | **0.636** | 0.641 | 0.594 | 0.521 | 0.464 |
| 2000 | | 0.669 | 0.600 | 0.632 | **0.672** | **0.636** | 0.533 | **0.467** |
| 5000 | | 0.670 | 0.595 | 0.611 | 0.662 | 0.579 | 0.533 | 0.463 |
| 10000 | | 0.668 | 0.587 | 0.596 | 0.651 | 0.554 | **0.546** | 0.463 |
| Average Change | | -0.033 | -0.061 | +0.006 | +0.076 | +0.018 | +0.009 | +0.103 |

The debiased version of the negative sentiment classifier was also tested based on the same contrastive clustering result, and produced similar results (Table 4). Any sentence containing annotations for anger, disgust, sadness, or fear was regarded as containing negative sentiment. The overall $F_1$ score is reduced slightly more by the debiased version of the classifier, and the reduction also applies to the sentences with bias propensities equal to or greater than 0.1. Otherwise, the debiased classifier provides higher $F_1$ scores for bias prone sentences, indicating it successfully reduces bias. The debiased classifier also performs significantly better on the sentences hand annotated as bias prone. The reduced effect size of the debiasing on the $F_1$ scores for negative sentiment may reflect the more generalized nature of negative sentiment: negative sentiment includes a variety of emotions, which may result in a wider range of topics appearing in the training data labeled with negative sentiment, which consequently lowers the risk that topic-specific distractor tokens will introduce bias.

**Per-Topic Impact of Bias on Anger and Negative Sentiment Prevalence Estimates**

Table 5: Estimated percentages of anger per topic compared to hand annotation. Includes the number of sentences N, the number annotated for anger, the percentages of sentences expressing anger according to the annotations (% Gold), original classifier (% Orig), and debiased classifier (% DeB), and the number of percentage points that the debiased classifier's estimate was closer to the annotated percentage than the original classifier's estimate (% Δ). The top three largest percent improvements are bolded.

| Topic Description | N | # Anger | % Gold | % Orig | % DeB | % Δ |
|---|---|---|---|---|---|---|
| General COVID-19 discussion | 6729 | 1767 | 26.3 | 46.8 | 17.2 | 11.5 |
| Masks, respirators, and gloves | 275 | 26 | 9.5 | 24 | 6.2 | 11.3 |
| Ventilators, lungs, and respirators | 243 | 27 | 11.1 | 19.8 | 4.9 | 2.5 |
| Politics and the vaccine | 233 | 84 | 36.1 | 58.4 | 15.5 | 1.7 |
| Virus origin discussions | 167 | 17 | 10.2 | 30.5 | 6 | 16.2 |
| "Downplaying" COVID-19 | 158 | 41 | 25.9 | 46.2 | 13.9 | 8.2 |
| Affected celebrities | 154 | 24 | 15.6 | 42.9 | 11.7 | 23.4 |
| Flattening the curve; hospital capacity | 123 | 8 | 6.5 | 8.1 | 1.6 | -3.3 |
| Taxes, funding, and tariffs | 108 | 20 | 18.5 | 51.9 | 6.5 | 21.3 |
| Democrats, reopening, and spread | 105 | 14 | 13.3 | 31.4 | 11.4 | 16.2 |
| Additional virus origin discussions | 102 | 49 | 48 | 90.2 | 36.3 | **30.4** |
| Progressives and voting | 99 | 11 | 11.1 | 26.2 | 5.1 | 9.1 |
| Schools and celebrities | 88 | 19 | 21.6 | 43.2 | 19.3 | 19.3 |
| Democrats, healthcare, and medicare | 87 | 11 | 12.6 | 31 | 5.7 | 11.5 |
| Claims of treatments or cures | 86 | 11 | 12.8 | 22.1 | 11.6 | 8.1 |
| Vaccines and immunity | 78 | 5 | 6.4 | 15.4 | 3.8 | 6.4 |
| Quarantine and food supply | 78 | 11 | 14.1 | 42.3 | 14.1 | **28.2** |
| School closures | 77 | 7 | 9.1 | 16.9 | 7.8 | 6.5 |
| The Constitution | 73 | 8 | 11 | 65.8 | 20.5 | **45.2** |
| Politicization | 65 | 29 | 44.6 | 78.5 | 21.5 | 10.8 |
| Racism | 41 | 21 | 51.2 | 97.6 | 80.5 | 17.1 |

The percentages of sentences for the topics that contain anger reveal significant variance across topics (Table 5). The original classifier's percentage estimate for anger always exceeds the hand annotated result, while the debiased classifier's estimate is almost always lower than the hand annotated result. However, the debiased classifier's estimate was closer to the hand annotated result for all but one topic. The top three topics exhibiting the largest improvement from debiasing demonstrate that certain issues can generate large inaccuracies in anger prevalence estimates. The generalized discussion ("none") topic had an estimated anger prevalence of 26.3%, but the original classifier's estimate was almost twice that: 46.8%. Even after debiasing the classifier, the estimate was 17.2%, which remains off by 9 percentage points.

As expected, given the smaller reduction in bias for negative sentiment according to $F_1$ scores, percentage changes for negative sentiment exhibit a similar but less pronounced pattern: while the average percent change for anger is 14.4%, it is 8.0% for negative sentiment.

## Discussion

Contrastive clustering significantly reduces the algorithmic bias present in emotion and sentiment classifier results on COVID-19 Reddit data, without requiring hand annotated data or classifier training. The algorithm slightly reduces the overall $F_1$ score of the classifiers in exchange for reducing the bias, as indicated by improving the $F_1$ scores for bias prone sentences. The bias reduction frequently overcompensates for bias in topic-level analyses, although the debiased estimates remain closer to the hand annotated estimates than the original classifier's estimates. Future research could examine whether these patterns hold on other data sets, and could explore mechanisms for calibrating the debiasing algorithm to reduce underestimates. Practitioners will need to determine whether the bias reduction is sufficiently important for their analyses to offset the reduction in the overall $F_1$ score. The bias reduction effect is significant for bias measured according to two independent methods: a bias propensity estimate for sentences based on hand annotations of individual tokens, and sentences hand annotated as prone to bias.

The contrastive clustering algorithm does not require hand annotating data. However, it requires setting several parameters, such as the relative importance of topic groups and emotion or sentiment groups in estimating which tokens express emotions, the threshold for determining which tokens express emotions or sentiments, and choosing how many tokens should be considered for masking. The threshold to designate a token as expressing an emotion or sentiment was tuned to achieve the most accurate contrastive clustering result. The remaining parameters worked well with their default values. Further testing is needed to determine whether these parameter values will generalize to other data sets and classification tasks.

There are multiple potential extensions of this approach. This study treated the classifiers as "black boxes" whose training data is unknown and cannot be retrained, which grants the algorithm broader practical applicability. However, the contrastive clustering approach could be incorporated into the classifier training. This could enable the neural networks to learn to reduce bias arising from conditionally dependent, context-dependent sets of tokens or phrases, rather than only reducing bias for individual tokens that are assumed to be conditionally independent. The black box approach could also be enhanced by considering masking sequences of tokens instead of single tokens, with the same goal of taking more context into account when choosing to mask tokens.

Future research could apply contrastive clustering to other types of bias by using different sets of clusters. For example, gendered words have been used to influence topic modeling algorithms to produce gender-specific topics, and then those gendered topics were manually contrasted with non-gendered topics to identify themes in gender bias [19]. Using gendered topics in contrastive clustering could extend the contrastive clustering algorithm to specifically address gender bias.

This could enable the contrastive clustering algorithm to replace the manual contrasting process, and enable further research on methods of constructing gender debiased classifiers or topic models.

This study used pre-trained emotion and sentiment classifiers whose training data would not have included COVID-19 discussions. The training data used to train these classifiers would impact the types of bias they learn, and how that bias would manifest for the COVID-19 data used in this study. The classifier accuracy should increase for COVID-19 for a custom trained classifier, since it would possess a more relevant vocabulary and examples of COVID-19 discussions in its training data. However, this could also increase a custom classifier's bias, since it could learn additional bias-prone tokens and phrases relating to COVID-19, such as organizations, people, treatments, and technologies.

Finally, quantifying bias is challenging. This study approaches bias as an overall algorithmic phenomenon, regardless of the underlying reasons for the bias. Multidisciplinary studies of the machine learning techniques for identifying algorithmic bias and social sciences approaches to categorizing bias would help reveal how effectively each type of bias can be mitigated with this approach, and suggest further lines of research to improve bias mitigation for each type of bias.

## Conclusion

Contrastive clustering provides an effective, unsupervised method of reducing classification bias in emotion and sentiment data for COVID-19 analyses. The technique slightly reduces the classification accuracy for sentences overall, in exchange for reducing bias. The bias reduction magnitudes for many discussion topics are large, suggesting that public health informatics studies should address potential bias when applying emotion and sentiment classifiers to social media data, either via bias reduction algorithms or hand annotation of sufficient samples of data. Further research is needed to determine how well the contrastive clustering technique generalizes to other classification tasks and data sets, and to explore possibilities to either mitigate the loss in overall accuracy or improve the bias reduction effect. Further research is also needed to explore what types of bias are reduced, and to evaluate the impact of the bias reduction on public health informatics analyses.

## References

1. Giustini DM, Ali SM, Fraser M and Boulos MNK: Effective uses of social media in public health and medicine: a systematic review of systematic reviews. Online Journal of Public Health Informatics. 2018;10. DOI: 10.5210/ojphi.v10i2.8270

2. Conway M, Hu M and Chapman WW: Recent advances in using Natural Language Processing to address public health research questions using Social Media and Consumer-Generated data. Yearb. Med. Inform.. 2019;28:208-217. DOI: 10.1055/s-0039-1677918

3. Medford RJ, Saleh SN, Sumarsono A, Perl TM and Lehmann CU: An "Infodemic": Leveraging High-Volume Twitter Data to Understand Early Public Sentiment for the Coronavirus Disease 2019 Outbreak. Open Forum Infectious Diseases. 2020;7. DOI: 10.1093/ofid/ofaa258

4. Saleh SN, Lehmann CU, McDonald SA, Basit MA and Medford RJ: Understanding public perception of coronavirus disease 2019 (COVID-19) social distancing on Twitter. Infection Control & Hospital Epidemiology. 2021;42:131–138. DOI: 10.1017/ice.2020.406

5. Wilson SL and Wiysonge C: Social media and vaccine hesitancy. BMJ Global Health. 2020;5. DOI: 10.1136/bmjgh-2020-004206

6. Valdez D, ten Thij M, Bathina K, Rutter LA and Bollen J: Social Media Insights Into US Mental Health During the COVID-19 Pandemic: Longitudinal Analysis of Twitter Data. J Med Internet Res. 2020;22:e21418. DOI: 10.2196/21418

7. Caton S and Haas C: Fairness in Machine Learning: A Survey. 2020. Available from: https://arxiv.org/abs/2010.04053

8. Rauchfleisch A and Kaiser J: The False positive problem of automatic bot detection in social science research. PLOS One. 2020;15:e0241045. DOI: 10.1371/journal.pone.0241045

9. Mowery J: Twitter Influenza Surveillance: Quantifying Seasonal Misdiagnosis Patterns. Online Journal of Public Health Informatics. 2016;8. DOI: 10.5210/ojphi.v8i3.7011

10. Zhang BH, Lemoine B and Mitchell M: Mitigating Unwanted Biases with Adversarial Learning. Proceedings of the 2018 AAAI/ACM Conference on AI, Ethics, and Society. 2018. DOI: 10.1145/3278721.3278779

11. Madras D, Creager E, Pitassi T and Zemel R: Learning Adversarially Fair and Transferable Representations. Proceedings of the 35th International Conference on Machine Learning. 2018;80:3384-3393. Available from: http://proceedings.mlr.press/v80/madras18a.html

12. Liu F and Avci B: Incorporating Priors with Feature Attribution on Text Classification. Proceedings of the 57th Annual Meeting of the Association for Computational Linguistics. 2019:6274-6283. DOI: 10.18653/v1/P19-1631

13. Ghili S, Kazemi E and Karbasi A: Eliminating Latent Discrimination: Train Then Mask. Proceedings of the AAAI Conference on Artificial Intelligence. 2019;33:3672-3680. DOI: 10.1609/aaai.v33i01.33013672

14. Caliskan A, Bryson JJ and Narayanan A: Semantics derived automatically from language corpora contain human-like biases. Science. 2017;356:183-186.

15. Bolukbasi T, Chang K-W, Zou J, Saligrama V and Kalai A: Man is to Computer Programmer as Woman is to Homemaker? Debiasing Word Embeddings. Proceedings of the 30th International Conference on Neural Information Processing Systems. 2016:4356–4364.

16. Sweeney C and Najafian M: Reducing Sentiment Polarity for Demographic Attributes in Word Embeddings Using Adversarial Learning. Proceedings of the 2020 Conference on Fairness, Accountability, and Transparency. 2020:359–368. DOI: 10.1145/3351095.3372837

17. Park JH, Shin J and Fung P: Reducing Gender Bias in Abusive Language Detection. Proceedings of the 2018 Conference on Empirical Methods in Natural Language Processing. 2018:2799-2804. DOI: 10.18653/v1/D18-1302

18. Romanov A, De-Arteaga M, Wallach H, Chayes J, Borgs C, Chouldechova A, Geyik S, Kenthapadi K, Rumshisky A and Kalai A: What's in a Name? Reducing Bias in Bios without Access to Protected Attributes. Proceedings of the 2019 Conference of the North American Chapter of the Association for Computational Linguistics: Human Language Technologies, Volume 1 (Long and Short Papers). 2019:4187-4195. DOI: 10.18653/v1/N19-1424

19. Devinney H, Björklund J and Björklund H: Semi-Supervised Topic Modeling for Gender Bias Discovery in English and Swedish. Proceedings of the Second Workshop on Gender Bias in Natural Language Processing. 2020:79-92. Available from: https://aclanthology.org/2020.gebnlp-1.8

20. Lahoti P, Beutel A, Chen J, Lee K, Prost F, Thain N, Wang X and Chi E: Fairness without Demographics through Adversarially Reweighted Learning. Advances in Neural Information Processing Systems. 2020;33:728-740. Available from: https://proceedings.neurips.cc/paper/2020/file/07fc15c9d169ee48573edd749d25945d-Paper.pdf

21. Hashimoto T, Srivastava M, Namkoong H and Liang P: Fairness Without Demographics in Repeated Loss Minimization. Proceedings of the 35th International Conference on Machine Learning. 2018;80:1929-1938. Available from: https://proceedings.mlr.press/v80/hashimoto18a.html

22. Liu EZ, Haghgoo B, Chen AS, Raghunathan A, Koh PW, Sagawa S, Liang P and Finn C: Just Train Twice: Improving Group Robustness without Training Group Information. Proceedings of the 38th International Conference on Machine Learning. 2021;139:6781-6792. Available from: https://proceedings.mlr.press/v139/liu21f.html

23. Kiritchenko S and Mohammad S: Examining Gender and Race Bias in Two Hundred Sentiment Analysis Systems. Proceedings of the Seventh Joint Conference on Lexical and Computational Semantics. 2018:43-53. DOI: 10.18653/v1/S18-2005

24. Barbieri F, Camacho-Collados J, Espinosa Anke L and Neves L: TweetEval: Unified Benchmark and Comparative Evaluation for Tweet Classification. Findings of the Association for Computational Linguistics: EMNLP 2020. 2020:1644-1650. DOI: 10.18653/v1/2020.findings-emnlp.148

25. Grootendorst M: BERTopic: Leveraging BERT and c-TF-IDF to create easily interpretable topics. 2020. DOI: 10.5281/zenodo.4381785

26. Newcombe RG: Two-sided confidence intervals for the single proportion: comparison of seven methods. Statistics in medicine. 1998;17:857-872.

27. Mikolov T, Grave E, Bojanowski P, Puhrsch C and Joulin A: Advances in Pre-Training Distributed Word Representations. Proceedings of the Eleventh International Conference on Language Resources and Evaluation (LREC 2018). 2018. Available from: https://aclanthology.org/L18-1008

28. Bernhardsson E: Approximate Nearest Neighbors Oh Yeah (ANNOY). 2013. Available from: https://github.com/spotify/annoy